\documentclass[runningheads]{llncs}
\usepackage{amsmath}
\usepackage{graphicx}
\usepackage{booktabs}
\usepackage{makecell}
\usepackage{wrapfig}
\usepackage{hyperref}
\usepackage{paralist}
\usepackage{tikz}
\usepackage{comment}
\usetikzlibrary{positioning}
\usetikzlibrary{shapes.geometric,arrows.meta,decorations.markings}
\usetikzlibrary{calc,shapes.callouts}

\makeatletter
\newcommand\scriptsizemy{\@setfontsize\scriptsize\@viiipt\@viiipt}
\makeatother

\makeatletter
\renewcommand\section{\@startsection{section}{1}{\z@}%
                       {-12\p@ \@plus -4\p@ \@minus -4\p@}%
                       {8\p@ \@plus 4\p@ \@minus 4\p@}%
                       {\normalfont\large\bfseries\boldmath
                        \rightskip=\z@ \@plus 8em\pretolerance=10000 }}
\makeatother

\begin{document}
\title{Guiding Inferences in Connection Tableau \\ by Recurrent Neural Networks
\vspace{-1mm}
}

\author{
Bartosz Piotrowski\inst{1}\inst{2}\thanks{
	Supported by the grant 2018/29/N/ST6/02903 of National Science Center, Poland.
	}
\and
Josef Urban\inst{1}\thanks{
	Supported by the \textit{AI4REASON} ERC Consolidator grant nr. 649043
	and by the Czech project AI\&Reasoning CZ.02.1.01/0.0/0.0/15\_003/0000466
	and the European Regional Development Fund.
	}
}

\institute{Czech Institute of Informatics, Robotics and Cybernetics, Prague,
Czech Republic
\and
Faculty of Mathematics, Informatics and Mechanics, University of Warsaw, Poland}
\authorrunning{Piotrowski \and Urban}

\maketitle
\begin{abstract}
\vspace{-2mm}
We present a dataset and experiments on applying recurrent neural networks
(RNNs) for guiding clause selection in the connection tableau proof calculus.
The RNN encodes a sequence of literals from the current branch of the partial
proof tree to a hidden vector state; using it, the system selects a clause for
extending the proof tree. The training data and learning setup are described,
and the results are discussed and compared with state of the art using gradient
boosted trees. Additionally, we perform a conjecturing experiment in which the
RNN does not just select an existing clause, but completely constructs the next
tableau goal.

\keywords{connection tableau \and neural networks \and internal guidance}

\end{abstract}

\section{Introduction}
\label{sec:intro}

There is a class of machine learning sequence-to-sequence architectures based
on recurrent neural networks (RNNs) which are successfully used in the domain
of natural language processing, in particular for translation between languages
\cite{Cho14}. Recently, such architectures proved useful also in various tasks
in the domain of symbolic computation
\cite{Wang18-short,Piotrowski19,Gauthier19,Lample19}. The models \emph{encode}
the source sequence to a \emph{hidden vector state} and \emph{decode} from it
the target sequence.

In this work, we employ such neural methods to choose among the
non-deterministic steps in connection-style theorem proving. In more detail, we
want to learn the \emph{hidden proving states} that correspond to the evolving
proof trees and condition the next prover steps based on them.  I.e., from a
set of connection tableau proofs we create a dataset (Section~\ref{sec:data})
of \textit{source-target} training examples of the form
(\textit{partial\_proof\_state, decision}) that we then use to train the neural
models (Section~\ref{sec:nmt_and_metric}). The results are reported in
Section~\ref{sec:results}. Section~\ref{sec:conj} shows an additional
experiment with predicting (conjecturing) tableau goals.

The connection tableau seems suitable for such methods. The connection proofs
grow as branches of a tree rooted in a starting clause. The number of options
(clauses) to choose from is relatively small compared to saturation-style
provers, where the number of clauses quickly grows to millions during the
search. The tableau branches representing the proof states can be the
sequential input to the RNNs, which can then decode one or more decisions,
i.e., choices of clauses.

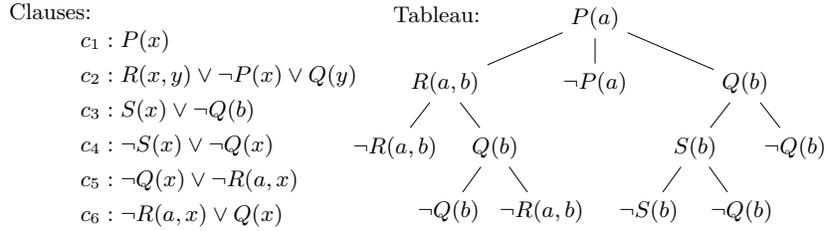
\begin{figure}
\begin{center}
\vspace{-1mm}
\scalebox{0.95}{
 \small
 \hspace{-15mm}
 \begin{minipage}{.35\textwidth}
 \hspace{-10mm}Clauses:\\
 $c_1 : P(x)$\\[1mm]
 $c_2 : R(x,y)\lor\lnot P(x) \lor Q(y)$\\[1mm]
 $c_3 : S(x)\lor\lnot Q(b)$\\[1mm]
 $c_4 : \lnot S(x)\lor \lnot Q(x)$\\[1mm]
 $c_5 : \lnot Q(x)\lor \lnot R(a,x)$\\[1mm]
 $c_6 : \lnot R(a,x) \lor Q(x)$
 \end{minipage}
 \raisebox{13.8mm}{Tableau:}\hspace{-20mm}
 \begin{minipage}{.35\textwidth}
\begin{tikzpicture}[level distance=9mm,
 level 1/.style ={sibling distance=2.1cm},
 level 2/.style ={sibling distance=1.4cm},
 level 3/.style ={sibling distance=1.3cm},
 ]
\node{$P(a)$}
 child { node {$R(a,b)$}
 child { node {$\lnot R(a,b)$} }
 child { node {$Q(b)$}
 child { node {$\lnot Q(b)$}}
 child { node {$\lnot R(a,b)$}}
 }
 }
 child { node {$\lnot P(a)$} }
 child { node {$Q(b)$}
 child { node {$S(b)$}
 child { node {$\lnot S(b)$}}
 child { node {$\lnot Q(b)$}}
 }
 child { node {$\lnot Q(b)$}}
 };
\end{tikzpicture}
\end{minipage}}
\vspace{-3mm}
\end{center}
\caption{
\label{tableau}
Closed connection tableau for a set of clauses.}
\end{figure}

\section{A Dataset for Connection-Style Internal Guidance}
\label{sec:data}

The experimental data originate from the Mizar Mathematical Library (MML)
\cite{Grabowski10} translated~\cite{Urban06} to the TPTP language. We have used
the leanCoP connection prover \cite{Otten03} to produce 13822 connection proofs
from the Mizar problems.

The connection tableau calculus searches for \emph{refutational proofs}, i.e.,
proofs showing that a set of first-order clauses is \emph{unsatisfiable}.
Figure~\ref{tableau} (adapted from Letz et al.~\cite{Letz94}) shows a set of
clauses and a \emph{closed connection tableau} constructed from them, proving
their unsatisfiability. A closed tableau is a tree with nodes labeled by
literals where each branch is \emph{closed}. A \emph{closed} branch contains a
pair of \emph{complementary literals} (with opposite polarities). An
\emph{open} branch can be extended with descendant nodes by applying one of the
input clauses. This \emph{extension} step can often be performed with several
different clauses -- this is the main non-determinism point. Choosing the
correct clause advances the proof, whereas choosing wrongly leads to redundant
exploration followed by backtracking.

The training data for choosing good clauses were extracted from the proofs as
follows.  First, formulas in the proofs were made more uniform by substituting
for each universal variable the token \texttt{VAR} and for each Skolem function
the token \texttt{SKLM}. For each non-root and non-leaf node $n$ in each proof
tree, we exported two types of paths, which form two kinds of input data for the
neural architecture:
\begin{compactenum}
	\item[(1)] $P_{\text{lits}}(r \rightarrow n)$ -- the literals
	leading from the root $r$ to the node $n$,
	\item[(2)] $P_{\text{cls}}(r \rightarrow n)$ -- the clauses that
		were chosen on the way from the root $r$ to $n$.
\end{compactenum}

The output data are created as follows. For each node $n$ we record the
decision (i.e., the clause) that led to the proof. Let $\text{clause}(n)$ be
the clause selected at node $n$.  For instance, if $n$ is the node labeled by
$R(a, b)$ in Figure \ref{tableau}, $\text{clause}(n) = c_6$.

The pairs $\big(P_{\text{lits}}(r \rightarrow n), \text{clause}(n)\big)$
and $\big(P_{\text{cls}}(r \rightarrow n), \text{clause}(n)\big)$ constitute
two different sets of training examples for learning clause selection. Each of
these sets contains 567273 pairs.  Additionally, we have constructed similar
data in which the output contains not only the choice of the next clause, but a
sequence of two or three such consecutive choices.
All these data sets\footnote{
The tableau proofs and the sequential training data extracted from it are
available at\\
\url{https://github.com/BartoszPiotrowski/guiding-connection-tableau-by-RNNs}}
were split into training, validation and testing sets -- the split was induced
by an initial split of the proofs in proportions 0.6, 0.1 and 0.3,
respectively.

%

\section{Neural Modelling and Evaluation Metric}
\label{sec:nmt_and_metric}

As a suitable sequence-to-sequence recurrent neural model we used an
implementation of a neural machine translation (NMT) architecture by Luong at
al. \cite{Luong17}, which was already successfully used for symbolic tasks in
\cite{Wang18-short} and \cite{Piotrowski19}.  All the hyperparameters used for
training were inherited from \cite{Wang18-short}.

Let $\textit{subsequent\_clauses}_i(P_\text{lits/cls}(r \rightarrow n))$ be a
set of $i$-long sequences of clauses found in the provided proofs, following a
given path of literals/clauses from the root to a node $n$. Note that since
there are alternative proofs in our data set, this set can have an arbitrary
size.  Let $\textit{clauses\_from\_model}^{\,i}_k(P_\text{lits/cls}(r \rightarrow
n))$ be a set of $k$ $i$-long sequences of clauses decoded from the NMT model
(we decoded for $k = 1$ or $k = 10$ most probable sequences using the
\emph{beam search}
technique \cite{Freitag17}). We consider the prediction from the model for a
given path of literals/clauses as successful if the sets
$\textit{subsequent\_clauses}_i(P_\text{lits/cls}(r \rightarrow n))$ and
$\textit{clauses\_from\_model}^{\,i}_k(P_\text{lits/cls}(r \rightarrow n))$ are
intersecting. The metric of predictive accuracy of the model is the proportion
of successful predictions on the test set.

\section{Results}
\label{sec:results}

\begin{wraptable}[13]{r}{67mm}
\vspace{-11mm}
\caption{\label{tab:results_inter_guid}
Predictive accuracy of the NMT system trained on two types of source paths
(literals or clauses), decoding 1-3 consecutive clauses. 1 or 10 best outputs
were decoded and assessed.
}
\centering
\vspace{2mm}
\begin{tabular}{c@{\hspace{3mm}}c@{\hspace{2mm}}c@{\hspace{3mm}}c@{\hspace{2mm}}c}
\toprule
& \multicolumn{2}{c}{\small{paths of literals}}
& \multicolumn{2}{c}{\small{paths of clauses}} \\
\cmidrule[0.07em](r){2-3} \cmidrule[0.07em](){4-5}
\thead{{\# clauses} \\ {to decode}}
& \thead{1 best\\output}
& \thead{10 best\\outputs}
& \thead{1 best\\output}
& \thead{10 best\\outputs} \\
\midrule
	1 & 0.64 & 0.72 & 0.17 & 0.36 \\
	2 & 0.11 & 0.19 & 0.03 & 0.07 \\
	3 & 0.05 & 0.07 & 0.01 & 0.02 \\
\bottomrule
\end{tabular}
\end{wraptable}


The average results for the above metric are shown in Table
\ref{tab:results_inter_guid}. We can see that predicting the next clause is
much more precise than predicting multiple clauses. The accuracy of predicting
the next clause(s) from a sequence of clauses is lower than predicting the
next clause(s) from a sequence of literals, which means the literals
give more precise information for making the correct decision.

We have also investigated how the performance of NMT depends on the length of
the input sequences. The results for the neural model trained on the paths of
literals as the input are shown in the second row of Table
\ref{tab:results_var_lengths}. As expected, the longer the input sequence, the
better is the prediction. The neural model was capable of taking advantage of a
more complex context. This differs significantly with the path-characterization
methods using manual features (as in~\cite{Kaliszyk18-short}) that just average
(possibly with some decay factor) over the features of all literals on the
path.

To compare with such methods, we trained a classifier based on gradient boosted
trees for this task using the XGBoost system~\cite{Chen16}, which was used for
learning feature-based guidance in~\cite{Kaliszyk18-short}. To make the task
comparable to the neural methods, we trained XGBoost in a multilabel setting,
i.e., for each partial proof state (a path of literals) it learns to score all
the available clauses, treated as labels. Due to limited resources, we restrict
this comparison to the MPTP2078 subset of MML which has 1383 different labels
(the clause names).

The average performance of XGBoost on predicting the next clause from the
(featurized) path of literals was 0.43. This is lower than the performance of
the neural model, also using literals on the path as the input (0.64). The
XGBoost performance conditioned on the length of the input path is shown in the
third row of Table~\ref{tab:results_var_lengths}. XGBoost is outperforming
NMT on shorter input sequences of literals, but on longer paths, XGBoost gets
significantly worse. The performance of the recurrent neural model grows with
the length of the input sequence, reaching 0.85 for input length 8. This means
that providing more context significantly helps the recurrent neural methods,
where the hidden state much more precisely represents (encodes) the whole path.
The feature-based representation used by XGBoost cannot reach such precision,
which is likely the main reason for its performance flattening early and
reaching at most 0.51.



\vspace{-4mm}
\begin{table}[h]
\begin{scriptsizemy}
\parbox{.49\linewidth}{
\caption{\label{tab:results_var_lengths}
Predictive accuracy of the NMT and XGBoost systems for different lengths of
input sequences consisting of literals.
}
\centering
\begin{tabular}{l@{\hspace{-0.2mm}}cccccccc}
\toprule
Length &    1 &    2 &    3 &    4 &    5 &    6 &    7 &    8 \\
\midrule
NMT & 0.19 & 0.48 & 0.64 & 0.70 & 0.68 & 0.72 & 0.79 & 0.85 \\
XGB & 0.43 & 0.35 & 0.42 & 0.39 & 0.47 & 0.41 & 0.51 & 0.46 \\
\bottomrule
\end{tabular}}
\hfill
\parbox{.48\linewidth}{
\centering
\vspace{-2.7mm}
\caption{\label{tab:conjecturing_literals}
Predictive accuracy of conjecturing literals by the NMT system for input
sequences of different lengths.
}
\begin{tabular}{l@{\hspace{-0.3mm}}cccccccc}
  \toprule
Length & 1 &    2 &    3 &    4 &    5 &    6 &    7 & \textbf{all}\\
\midrule
NMT & 0.04 & 0.05 & 0.08 & 0.11 & 0.14 & 0.16 & 0.34 & 0.08 \\
\bottomrule
\end{tabular}
}
\end{scriptsizemy}
\end{table}
\vspace{-7mm}

\section{Conjecturing New Literals}
\label{sec:conj}
As an additional experiment demonstrating the power of the recurrent neural
methods we constructed a data set  for \emph{conjecturing} new literals on the
paths in the tableau proofs. The goal here is not to select a proper literal,
but to \textit{construct} it from the available symbols (the number of them for
the MML-based data set is 6442). This task is impossible to achieve with the
previous methods that can only \emph{rank} or \emph{classify} the available
options. Recurrent neural networks are, on the other hand, well-suited for such
tasks -- e.g., in machine translation, they can learn how to compose
grammatically correct and meaningful sentences.

It turns out that this more difficult task is to some extent feasible with NMT.
Table \ref{tab:conjecturing_literals} shows that NMT could propose the right
next literal on the path in a significant number of cases. Again, there is a
positive dependence between the length of the input sequence and the predictive
performance. Most of the times the correct predictions involve short literals,
whereas predicting longer literals is harder. The proposed longer literals
often not only do not match the right ones but have an improper structure (see
Table \ref{tab:conjecturing_examples} for examples of the NMT outputs).

\vspace{-3.5mm}
\begin{table}
\caption{\label{tab:conjecturing_examples}
Literals conjectured by NMT \textit{vs.} the correct ones. (1) is an example of a
correctly predicted output; in (2) NMT was wrong but proposed a literal which
is similar to the proper one; (3) shows a syntactically incorrect literal
produced by NMT.}
\centering
\vspace{-1mm}
\scalebox{0.90}{
\begin{tabular}{l@{\hspace{5pt}}l@{\hspace{10pt}}l}
\toprule
& \textbf{NMT prediction} & \textbf{Correct output} \\
\midrule
	(1) & \texttt{m1\_subset\_1(np\_\_1,k4\_ordinal1)}&\texttt{m1\_subset\_1(np\_\_1,k4\_ordinal1)}\\
	(2) & \texttt{m1\_subset\_1(SKLM,k1\_zfmisc\_1(SKLM))}&\texttt{m1\_subset\_1(SKLM,SKLM)}\\
	(3) & \texttt{k2\_tarski(SKLM,SKLM)=k2\_tarski(SKLM}&\texttt{k2\_tarski(SKLM,SKLM)=k2\_tarski(SKLM,SKLM)}\\
	\bottomrule
\end{tabular}}
\vspace{-2mm}
\end{table}

\section{Conclusion and Future Work}

In this work, we proposed RNN-based encoding and decoding as a suitable
representation and approach for learning clause selection in connection
tableau. This differs from previous approaches -- both neural and non-neural --
by emphasizing the importance of the evolving proof state and its accurate
encoding. The approach and the constructed datasets also allow us to
meaningfully try completely new tasks, such as automatically conjecturing the
next literal on the path.  The experimental evaluation is encouraging.  In
particular, it shows that the longer the context, the more precise the
recurrent methods are in choosing the next steps, unlike the previous methods.
The evaluation and data sets have focused (as similar research
studies~\cite{Kaliszyk17-short,Evans18-short}) on the machine learning
performance, which is known to underlie the theorem proving performance. Future
work includes integrating such methods into ATP systems and ATP evaluation
similar to~\cite{Kaliszyk18-short,ChvalovskyJ0U19-short}.

\bibliographystyle{splncs04}
\bibliography{references}

\begin{thebibliography}{10}
\providecommand{\url}[1]{\texttt{#1}}
\providecommand{\urlprefix}{URL }
\providecommand{\doi}[1]{https://doi.org/#1}

\bibitem{Chen16}
Chen, T., Guestrin, C.: XGBoost: {A} scalable tree boosting system. In: {ACM}
  {SIGKDD} 2016, pp. 785--794 (2016)

\bibitem{Cho14}
Cho, K., van Merrienboer, B., G{\"{u}}l{\c{c}}ehre, {\c{C}}., Bahdanau, D.,
  Bougares, F., Schwenk, H., Bengio, Y.: Learning phrase representations using
  {RNN} encoder-decoder for statistical machine translation. In: {EMNLP} 2014,
  pp. 1724--1734 (2014)

\bibitem{ChvalovskyJ0U19-short}
Chvalovsk{\'{y}}, K., Jakubuv, J., Suda, M., Urban, J.: {ENIGMA-NG:}
  {E}fficient neural and gradient-boosted inference guidance for {E}. In:
  {CADE} 27, pp. 197--215 (2019)

\bibitem{Evans18-short}
Evans, R., Saxton, D., Amos, D., Kohli, P., Grefenstette, E.: Can neural
  networks understand logical entailment? In: {ICLR} 2018 (2018)

\bibitem{Freitag17}
Freitag, M., Al{-}Onaizan, Y.: Beam search strategies for neural machine
  translation. In: NMT@ACL 2017, pp. 56--60 (2017)

\bibitem{Gauthier19}
Gauthier, T.: Deep reinforcement learning for synthesizing functions in
  higher-order logic. CoRR  (2019), \url{http://arxiv.org/abs/1910.11797}

\bibitem{Grabowski10}
Grabowski, A., Kornilowicz, A., Naumowicz, A.: Mizar in a nutshell. J.
  Formalized Reasoning  \textbf{3}(2),  153--245 (2010)

\bibitem{Kaliszyk17-short}
Kaliszyk, C., Chollet, F., Szegedy, C.: Holstep: {A} machine learning dataset
  for higher-order logic theorem proving. In: {ICLR} 2017 (2017)

\bibitem{Kaliszyk18-short}
Kaliszyk, C., Urban, J., Michalewski, H., Ols{\'{a}}k, M.: Reinforcement
  learning of theorem proving. In: NeurIPS 2018, pp. 8836--8847 (2018)

\bibitem{Lample19}
Lample, G., Charton, F.: Deep learning for symbolic mathematics. CoRR  (2019),
  \url{http://arxiv.org/abs/1912.01412}

\bibitem{Letz94}
Letz, R., Mayr, K., Goller, C.: Controlled integration of the cut rule into
  connection tableau calculi. J. Autom. Reasoning  \textbf{13},  297--337
  (1994)

\bibitem{Luong17}
Luong, M., Brevdo, E., Zhao, R.: Neural machine translation (seq2seq) tutorial
  (2017), \url{https://github.com/tensorflow/nmt}

\bibitem{Otten03}
Otten, J., Bibel, W.: {leanCoP: L}ean connection-based theorem proving. J.
  Symbolic Computation  \textbf{36}(1-2),  139--161 (2003)

\bibitem{Piotrowski19}
Piotrowski, B., Urban, J., Brown, C.E., Kaliszyk, C.: Can neural networks learn
  symbolic rewriting? CoRR  (2019), \url{https://arxiv.org/pdf/1911.04873.pdf}

\bibitem{Urban06}
Urban, J.: {MPTP} 0.2: {D}esign, implementation, and initial experiments. J.
  Automated Reasoning  \textbf{37}(1-2),  21--43 (2006)

\bibitem{Wang18-short}
Wang, Q., Kaliszyk, C., Urban, J.: First experiments with neural translation of
  informal to formal mathematics. In: {CICM} 2018, pp. 255--270 (2018)

\end{thebibliography}

\end{document}